\documentclass{article}
\usepackage{spconf,amsmath,graphicx,amssymb,booktabs,caption,subcaption,array,threeparttable,stfloats,bbding,color,enumitem}
\title{EvoPose: a Recursive transformer for 3D Human Pose Estimation with kinematic structure priors}
\name{Yaqi Zhang$^{1,2}$, Yan Lu$^{3}$, Bin Liu$^{1,2\star}$\thanks{$^{\star}$Corresponding author, email: flowice@ustc.edu.cn.}, Zhiwei Zhao$^{1,2}$, Qi Chu$^{1,2}$, Nenghai Yu$^{1,2}$}
\address{$^{1}$School of Cyber Science and Technology, University of Science and Technology of China \\
$^{2}$CAS Key Laboratory of Electromagnetic Space Information\quad$^{3}$The University of Sydney}
\begin{document}
%
\maketitle
\let\thefootnote\relax\footnotetext{This work is supported by the National Natural Science Foundation of China (Grant No. 62272430).}
\begin{abstract}
Transformer is popular in recent 3D human pose estimation, which utilizes long-term modeling to lift 2D keypoints into the 3D space. 
However, current transformer-based methods do not
fully exploit the prior knowledge of the human skeleton provided by the kinematic structure. 
In this paper, we propose a novel transformer-based model EvoPose to introduce the human body prior knowledge for 3D human pose estimation effectively. 
Specifically, a Structural Priors Representation (SPR) module represents human priors as structural features carrying rich body patterns, e.g. joint relationships. The structural features are interacted with 2D pose sequences and help the model to achieve more informative spatiotemporal features.
Moreover,
a Recursive Refinement (RR) module is applied to refine the 3D pose outputs by utilizing estimated results and further injects human priors simultaneously. 
Extensive experiments demonstrate the effectiveness of EvoPose which achieves a new state of the art on two most popular benchmarks, Human3.6M and MPI-INF-3DHP.
\end{abstract}
\section{Introduction}
\label{sec:intro}
Monocular 3D human pose estimation (HPE) aims to estimate 3D joint positions of the human skeleton from given videos, which has rich practical application scenarios, such as motion capture~\cite{desmarais2021review,li2022lidarcap}
and virtual 
reality~\cite{mehta2017vnect}. An effective pipeline is separating the 3D HPE as a two-stage system consisting of 2D keypoints detection and 3D joints lifting~\cite{chen20173d}. 
Between them, the key barrier is the 3D joints lifting because of the ill-posed property caused by the depth leakage. 
Existing state-of-the-art 3D HPE methods~\cite{li2022exploiting,zheng20213d,hassanin2022crossformer} solve the lifting problem by modeling long-term and fine-grained spatiotemporal information with Transformer~\cite{vaswani2017attention} and achieve advanced results. But they often merely extract features for 2D pose sequences, which misses the human priors and limits their performances.

We believe that human priors, also called kinematic structural information, can provide much more useful information to constrain the estimated human poses, which has the potential to lead to more reliable results. In particular, kinematic cues of human skeleton~\cite{kundu2020kinematic} provide body structural knowledge that indicates the category of each joint and the connectivities~\cite{kundu2020kinematic} for every joint pair, which results in more plausible 3D poses in physical. To allocate that prior knowledge more effectively in the Transformer model for 3D HPE, we propose a novel transformer lifting model EvoPose.

EvoPose includes a Structural Priors Representation (SPR) module, a SpatioTemporal Enhancement (STE) module, and a Recursive Refinement (RR) module to introduce human priors effectively. Firstly, the SPR extracts structural features from the kinematic tree to introduce human priors initially. And then in the STE module, the 2D pose sequence is combined with the structural features in a STEvo block to inject the kinematic constraint in the spatiotemporal modeling process, leading to more informative spatiotemporal sequence features. Finally, the RR module estimates 3D poses based on these sequence and structural features recursively to further incorporate the human priors. With the aforementioned processes, the human priors have been introduced to 3D HPE step-by-step. 
Our main contributions are as follows: 
\begin{itemize}[itemsep=0pt,topsep=0pt,parsep=0pt]
\item We present a novel transformer method, EvoPose, for monocular 3D HPE using three modules to introduce kinematic structure priors effectively step-by-step.
\item Our EvoPose achieves state-of-the-art performance on two datasets, surpassing existing methods by 10.3mm under MPJPE on MPI-INF-3DHP especially.
\end{itemize}

\section{Methods}
\begin{figure*}[ht]
  \setlength{\abovecaptionskip}{0cm}
  \setlength{\belowcaptionskip}{-0.3cm}
  \centering
  \centerline{\includegraphics[width=.85\textwidth,trim=6 12 35 6.6,clip]{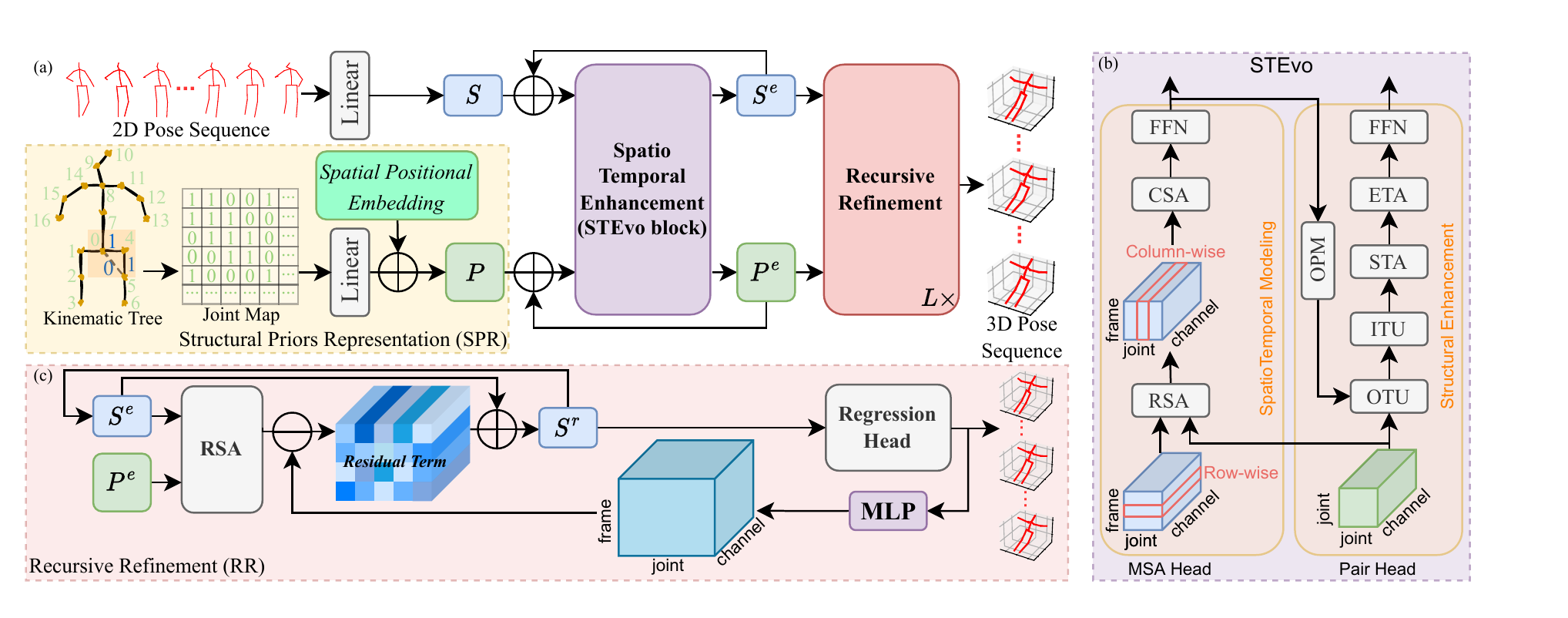}}
\caption{{\bf (a)} Overview of the proposed EvoPose. Structural Priors Representation (SPR) derives structural features from the kinematic tree. SpatioTemporal Enhancement (STE) models spatiotemporal relations and enhances structural features. {\bf (b)} Overview of STEvo in STE. {\bf (c)} Overview of the Recursive Refinement (RR) which refines the estimated result auto-regressively.}
\label{fig:model}
\end{figure*}

The overview of our proposed EvoPose is illustrated in Fig.~\ref{fig:model}. Firstly, 
a Structural Priors Representation (SPR) module formulates
the kinematic tree as structural features $P$ to indicate the high-level relations in each joint pair. Then, a transformer module SpatioTemporal Enhancement (STE) guides $S$ and $P$ to interact with each other and extracts stronger sequence and structural features (defined as $S^e$ and $P^e$) 
for further usage. Finally, a Recursive Refinement (RR) module estimates 3D poses by modeling on $S^e$ and $P^e$ recursively.

\subsection{Structural Priors Representation}
To introduce the human priors into our model, the SPR first builds a joint map $M\in\mathbb R^{J\times J\times 1}$ as Fig.~\ref{fig:model}(a) shows and $J$ is the total number of body joints. Specifically, if joint $i$ and $j$ are connected in the kinematic tree, then $M_{ij}=M_{ji}=1$, otherwise $M_{ij}=M_{ji}=0$. This map represents joints connection in the kinematic tree, indicating the local relationships in the human body. We further use a Spatial Positional Embedding (SPE) to introduce the global kinematic cues. Because the ordering of joints follows the rules of the kinematic tree, the gap of joint number can reflect the category difference in each joint pair, providing more global relationships. We formulate these gaps into one-hot vectors 
and map them to SPE as a tensor with the shape of $J\times J\times d_p$, where $d_p$ is the structural feature dimension. After that, the joint map is fed into a linear layer to expand its dimension and added with the SPE to get the structural features $P\in\mathbb R^{J\times J\times d_p}$.
\subsection{SpatioTemporal Enhancement}
When the SPR processes human body priors, the 2D pose sequence is also transferred to initial features $S\in\mathbb R^{N\times J\times d_s}$, where $N$ is the frame number and $d_s$ is the sequence feature dimension. 
To integrate human priors into the spatiotemporal modeling process, we propose a SpatioTemporal Evoformer (STEvo) block into the STE module to incorporate the human priors $P$ and the sequence features $S$ effectively.\vspace{-0.2em}
\subsubsection{SpatioTemporal Evoformer}
The original Evoformer~\cite{jumper2021highly} has two input heads, a multiple sequence alignment (MSA) representation one, and a pair representation one. In the STEvo, we send the temporal features $S$ to the MSA head and the human structural features $P$ to the pair representation head, which is shown in Fig.~\ref{fig:model}(b). Obviously, our inputs have quite different meanings from the original Evoformer, but they share the same dimension formats. Specifically, 
the pose features $S$, sent to the MSA head, have the shape of $N\times J\times d_s$, and the structural features $P$, fed to the pair representation head, possess $J\times J\times d_p$ shape in our STEvo. The message interactions of $S$ in STEvo can be interpreted as two levels: the spatial level on $J$ joints and the temporal level across $N$ frames. In this process, human priors provided by $P$ can be injected into the pose features $S$ effectively and $P$ are also interacted with $S$, leading to more expressive spatiotemporal features $S^e$ and further enhanced structural features $P^e$.\vspace{-0.2em}

\subsubsection{Details of SpatioTemporal Evoformer}
In this subsection, we give further insights and details of STEvo. Its details are shown in the Fig.~\ref{fig:model}(b). We denote the branch processing $S$ as the SpatioTemporal Modeling branch. And the branch for $P$ is the Structural Enhancement branch. 

\noindent{\bf SpatioTemporal Modeling.} Following the original Evoformer, our STEvo also uses row- and column-wise self-attention (RSA and CSA)~\cite{jumper2021highly} to treat the MSA head input--pose features $S$. The RSA means the self-attention for each row of $S$. Note that $S$ has the shape of $N\times J\times d_s$ and its row is a $J\times d_s$ matrix which means the human pose features in one frame. So the RSA is essential to the spatial message interaction and can be written as follows:
\setlength\abovedisplayskip{0.1em}
\setlength\belowdisplayskip{0.1em}
\begin{equation}
    Attn_i = \text{Softmax}(Q_iK_i^T/\sqrt{d_s}+\text{Linear}(P))V_i,\label{attention}
\end{equation}
where $Q_i\in\mathbb R^{J\times d_s}$, $K_i\in\mathbb R^{J\times d_s}$ and $V_i\in\mathbb R^{J\times d_s}$ denote query, key and value~\cite{vaswani2017attention} respectively, which are computed by pose features $S_i\in\mathbb R^{J\times d_s}$ of the $i$-th frame. Compared to the vanilla transformer~\cite{vaswani2017attention}, RSA injects human priors $P$ into the affinities to constrain relationships between different joints, which makes the spatial information interaction more reliable. The CSA is the vanilla self-attention for each column ($\mathbb R^{N\times d_s}$) of $S$, which is obviously corresponding to the temporal interaction for each joint. So the cascade of RSA and CSA can model the spatiotemporal patterns with human kinematic priors and lead to enhanced pose features $S^e$.


\noindent{\bf Structural Enhancement.} The structural features $P$ are updated by the enhanced pose features $S^e$. The outer product mean (OPM)~\cite{jumper2021highly} is first applied on different rows of $S^e$:
\begin{equation}\label{ffn}
    F_{ij} = \frac{1}{N}\sum_{n=0}^{N-1} S^e_{n,i}\otimes S^e_{n,j},
\end{equation}
where $S^e_{n, i}\in\mathbb R^{N\times d_s}$ denotes the features of the $i$-th joint at the $n$-th frame. The output $F_{ij}\in\mathbb R^{d_s\times d_s}$ means the temporal averaged outer product matrix along the whole $N$ frames, indicating the high-order features of the relationship between the $i$-th and $j$-th joints. And then, these high-order features $F\in\mathbb R^{J\times J\times d_s\times d_s}$ are used to update $P$ as follows:
\begin{equation}
    \widetilde P = P + \text{Linear}(F).
\end{equation}
This $\text{Linear}(\cdot)$ is a fully-connect layer operated on the channel level: $\mathbb R^{d_s\times d_s}\rightarrow\mathbb R^{d_p}$, which makes the dimension of $F$ matched with $P$. After that, several non-attention and attention modules, including triangular updates using outgoing and incoming edges (OTU and ITU) and triangular self-attention around starting and ending node (STA and ETA)~\cite{jumper2021highly} which are utilized in the original Evoformer, are operated on $\widetilde P$ to get the final refined $P^e$. These refined structural features incorporate two kinds of human priors. One is extracted by the kinematic tree and the other is derived from the given pose sequence, leading to stronger human pose prior knowledge and improving the final 3D pose estimation accuracy.\vspace{-0.3em}
\subsection{Recursive Refinement}\vspace{-0.2em}
After obtaining the enhanced spatiotemporal and structural features $S^e$ and $P^e$, the Recursive Refinement module estimates the 3D poses by the following recursive pipeline:
\begin{equation}
    S_t^r = \text{FeatRe}(S_t^e, X_{t-1}),\ \ X_t = \text{RegHead}(S_t^r), 
\end{equation}
where $X_t$ is the 3D pose estimation results at $t$-th round and $S_t^e=S_{t-1}^r$. Note that, $S^e_1=S^e$, $X_0=\bf{0}$, and the pipeline is shown in Fig.~\ref{fig:model}(c). The \text{FeatRe} refines the pose features $S^e$ by introducing the last round pose estimation results $X_{t-1}$:
\begin{equation}
\begin{aligned}
    S_t^r &= \text{FeatRe}(S_t^e, X_{t-1})\\
          &= \underbrace{\text{RSA}(S_t^e, P^e) - \text{MLP}(X_{t-1})}_{\text{residual term}} + S_{t}^e.
\end{aligned}
\end{equation}
The residual term estimates the corresponding gap between $S_t^e$ and $X_{t-1}$, reflecting new information that the $X_{t-1}$ lack but $S_t^e$ provide. The $\text{MLP}(\cdot)$ means a 3 layer multi-layer perception.  This new information is added on the $S_{t}^e$ to get the refined one $S_t^r$. The $S_t^r$ has the shape of $N\times J \times d_s$, expressing each joint as a $d_s$-dimensional vector. So the RegHead is a joint-level convolutional module that maps each joint to its 3D coordinates. The last round estimation results are defined as $X^{3d}\in\mathbb R^{N\times J\times 3}$, storing the $(x,y,z)$ coordinate value for each joint across all frames. The recursive estimation pipeline further injects the human priors in the final 3D pose estimation, leading to more reliable results. 

\vspace{-0.5em}

\subsection{Loss Function}
The model is trained end-to-end with several types of loss functions. 
We first compute a coordinate loss $\mathcal L_c$ by the Mean Squared Error between estimated and ground truth 3D poses coordinates to constrain spatial error.
Additionally, we compute the joint velocity loss~\cite{pavllo20193d} $\mathcal L_v$ and acceleration loss~\cite{xu20213d} $\mathcal L_a$, which measure the gap of the first-order and second-order derivative of the 3D pose coordinates, respectively. 
Moreover, we also project predicted 3D poses to 2D and compute the reprojection error between them and the ground truth 2D pose as the reprojection loss $\mathcal L_p$. 
Finally, these losses are combined together into a total one: $\mathcal L = \mathcal L_c + \lambda_v\cdot\mathcal L_v + \lambda_a\cdot\mathcal L_a+\lambda_p\cdot\mathcal L_p$, where $\lambda_\bullet$ is the weight of each loss term.
\begin{table*}[ht]
    \setlength{\abovecaptionskip}{0cm}
    \setlength{\belowcaptionskip}{-0.3cm}
    \caption{Results comparison with state-of-the-art methods on Human3.6M under MPJPE(mm). Top: 2D poses detected by CPN; Bottom: ground truth 2D poses. Bold: the best; Underline: the second.}
    \label{tab:h36m}
    \centering
    \resizebox{\textwidth}{!}{
    \begin{tabular}{@{}l|ccccccccccccccc|c@{}}
        \toprule
        Method & Dir. & Disc & Eat & Greet & Phone & Photo & Pose & Purch. & Sit & SitD. & Smoke & Wait & WalkD. & Walk & WalkT. & Avg. \\
        \midrule
        PoseFormer (ICCV'21)\cite{zheng20213d} & 41.5 & 44.8 & \underline{39.8} & 42.5 & 46.5 & 51.6 & 42.1 & 42.0 & \underline{53.3} & 60.7 & 45.5 & 43.3 & 46.1 & 31.8 & 32.2 & 44.3 \\
        MHFormer (CVPR'22)\cite{li2022mhformer} & \underline{39.2} & 43.1 & 40.1 & \underline{40.9} & \underline{44.9} & \underline{51.2} & {\bf 40.6} & \underline{41.3} & 53.5 & 60.3 & {\bf 43.7} & {\bf 41.1} & 43.8 & 29.8 & 30.6 & 43.0 \\
        P-STMO (ECCV'22)\cite{shan2022p} & {\bf 38.9} & {\bf 42.7} & 40.4 & 41.1 & 45.6 & {\bf 49.7} & \underline{40.9} & {\bf 39.9} & 55.5 & \underline{59.4} & \underline{44.9} & 42.2 & {\bf 42.7} & \underline{29.4} & {\bf 29.4} & \underline{42.88} \\
        \midrule
        EvoPose (Ours) & 41.7 & \underline{43.0} & {\bf 38.1} & {\bf 40.7} & {\bf 44.2} & 52.5 & 41.3 & 42.6 & {\bf 52.7} & {\bf 56.8} & 45.3 & \underline{41.5} & \underline{42.9} & {\bf 28.8} & \underline{29.6} & {\bf 42.83} \\
        \bottomrule
        \toprule
        Method & Dir. & Disc & Eat & Greet & Phone & Photo & Pose & Purch. & Sit & SitD. & Smoke & Wait & WalkD. & Walk & WalkT. & Avg. \\
        \midrule
        PoseFormer (ICCV'21)\cite{zheng20213d} & 30.0 & 33.6 & 29.9 & 31.0 & 30.2 & 33.3 & 34.8 & 31.4 & 37.8 & 38.6 & 31.7 & 31.5 & 29.0 & 23.3 & 23.1 & 31.3 \\
        MHFormer (CVPR'22)\cite{li2022mhformer} & \underline{27.7} & 32.1 & 29.1 & 28.9 & 30.0 & 33.9 & 33.0 & 31.2 & 37.0 & 39.3 & 30.0 & 31.0 & 29.4 & 22.2 & 23.0 & 30.5 \\
        P-STMO (ECCV'22)\cite{shan2022p} & 28.5 & \underline{30.1} & \underline{28.6} & \underline{27.9} & \underline{29.8} & \underline{33.2} & \underline{31.3} & \underline{27.8} & \underline{36.0} & \underline{37.4} & \underline{29.7} & \underline{29.5} & \underline{28.1} & \underline{21.0} & \underline{21.0} & \underline{29.3} \\
        \midrule
        EvoPose (Ours) & {\bf 24.3} & {\bf 24.8} & {\bf 23.1} & {\bf 23.4} & {\bf 24.6} & {\bf 25.9} & {\bf 28.3} & {\bf 24.6} & {\bf 30.1} & {\bf 31.4} & {\bf 25.3} & {\bf 24.1} & {\bf 23.7} & {\bf 18.7} & {\bf 20.1} & {\bf 24.8} \\
        \bottomrule
    \end{tabular}
    }
\end{table*}
\begin{table}[t]\footnotesize
    \setlength{\abovecaptionskip}{0cm}
    \setlength{\belowcaptionskip}{-0.3cm}
    \caption{Results comparison with state-of-the-art methods on MPI-INF-3DHP. Bold: the best; Underline: the second.}
    \label{tab:3dhp}
    \centering
    \begin{tabular}{@{}l|c|ccc@{}}
        \toprule
        Methods & N & PCK$\uparrow$ & AUC$\uparrow$ & MPJPE$\downarrow$ \\
        \midrule
        PoseFormer (ICCV'21)\cite{zheng20213d} & 9 & 88.6 & 56.4 & 77.1 \\
        MHFormer (CVPR'22)\cite{li2022mhformer} & 9 & 93.8 & 63.3 & 58.0 \\
        P-STMO (ECCV'22)\cite{shan2022p} & 81 & {\bf 97.9} & 75.8 & 32.2 \\
        \midrule
        EvoPose (Ours) & 9 & \underline{97.8} & \underline{81.9} & \underline{24.2} \\
        EvoPose (Ours) & 27 & \underline{97.8} & {\bf 83.7} & {\bf 21.9} \\
        \bottomrule
    \end{tabular}
\end{table}
\begin{table}[t]\footnotesize
    \setlength{\abovecaptionskip}{0cm}
    \setlength{\belowcaptionskip}{-0.3cm}
    \caption{Ablation study on model components: SPR and the recursive pipeline in RR.}
    \label{tab:ablation}
    \centering
    \begin{tabular}{cc|c}
        \toprule
        SPR & Recursive Pipeline (RR) & MPJPE \\
        \midrule
        \XSolidBrush & \XSolidBrush & 57.3 \\
        \Checkmark & \XSolidBrush & 47.9 \\
        \XSolidBrush & \Checkmark & 55.5 \\
        \Checkmark & \Checkmark & {\bf 45.8} \\
        \bottomrule
    \end{tabular}
\end{table}
\begin{figure*}[t]
  \setlength{\abovecaptionskip}{0cm}
  \setlength{\belowcaptionskip}{-0.2cm}
  \centering
  \centerline{\includegraphics[width=\textwidth,trim=0 15 14.4 8.8,clip]{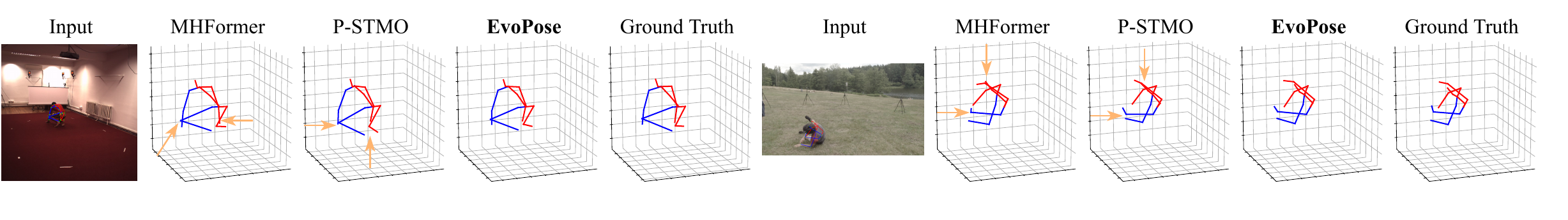}}
\caption{Qualitative comparison with two state-of-the-art methods on both Human3.6M (left) and MPI-INF-3DHP (right) dataset.}
\label{fig:vis}
\end{figure*}

\section{Experiments}
\label{sec:experiments}
{\bf Implementation details.} 
In our implementation, we set $\lambda_v = \lambda_a = 0.2$ and $\lambda_p = 0.1$.  The RR module is conducted for $L=2$ times. The proposed EvoPose is implemented by PyTorch on Tesla V100 GPU. We use Amsgrad optimizer with an initial learning rate of 0.001 which decays by 5\% after each epoch and 50\% after every 5 epochs. 
For fair comparisons, we also adopt horizontal flip augmentation following~\cite{zheng20213d,pavllo20193d,chen2021anatomy}. Note that, we only use the center frame of the final results as the prediction in inference.

\subsection{Datasets and Evaluation Metrics}
\noindent{\bf Human3.6M.} The Human3.6M~\cite{h36m_pami} is the largest indoor dataset for 3D HPE, containing 3.6 million video frames with 15 scenarios recorded by four calibrated cameras. Following~\cite{chen2021anatomy,zheng20213d,li2022mhformer,shan2022p}, we train our model on five subjects and test on two subjects with 17 joints. We report the mean per joint position error (MPJPE).

\noindent{\bf MPI-INF-3DHP.} The MPI-INF-3DHP~\cite{mono-3dhp2017} is a large challenging 3D pose dataset with both indoor and outdoor scenarios. 
Following the settings in previous works~\cite{zheng20213d,li2022mhformer,shan2022p}, we use eight subjects for training and six subjects for testing on valid frames captured by eight cameras. We use three evaluation metrics: MPJPE, percentage of correct keypoints (PCK) under the threshold of 150mm, and area under curve (AUC).
\begin{table}[t]\footnotesize
    \setlength{\abovecaptionskip}{0cm}
    \setlength{\belowcaptionskip}{-0.3cm}
    \caption{Ablation study on the number of input frames with MPJPE. CPN: cascaded pyramid network; GT: ground truth.}
    \label{tab:receptive}
    \centering
    \begin{tabular}{ccccc}
        \toprule
        2D Inputs & 9 & 27 & 81 & 243 \\
        \midrule
        CPN & 48.3 & 45.8 & 43.7 & {\bf 42.8} \\
        GT & 34.6 & 33.1 & 28.0 & {\bf 24.8} \\
        \bottomrule
    \end{tabular}
\end{table}

\subsection{ Comparison with State-of-the-Art Methods}
{\bf Results on Human3.6M.}
We compare our model with state-of-the-art methods. Cascaded pyramid network (CPN)~\cite{chen2018cascaded} is used to estimate 2D poses from video frames, following~\cite{zheng20213d,li2022mhformer,shan2022p,pavllo20193d}. We report the results in Table~\ref{tab:h36m} (top) with inputs of 243 frames and achieve {\bf 42.83mm} under MPJPE. It can be seen that our EvoPose is competitive to P-STMO~\cite{shan2022p}. To further explore the lower bound of our method, the results with ground truth 2D poses are reported in Table~\ref{tab:h36m} (bottom). EvoPose surpasses all the other methods by a large margin, demonstrating the effectiveness of our method. When the input 2D poses are more accurate to be consistent with human kinematic, the performance of our method becomes better.

\noindent{\bf Results on MPI-INF-3DHP.}
To evaluate the ability of our model in the real world, we also report the performance on MPI-INF-3DHP compared to state-of-the-art methods. Following~\cite{chen2021anatomy,li2022mhformer,shan2022p}, we use ground truth 2D poses as inputs. Due to the sequence lengths of this dataset being shorter than Human3.6M, we adopt the setting of 27 frames. The results are shown in Table~\ref{tab:3dhp}. It can be seen that our method performs much well compared to all the other methods with improvements of {\bf 10.3mm} in MPJPE and {\bf 7.9} in AUC over the previous best method, P-STMO, and achieves competitive performance in PCK with it by using smaller input length. This indicates the great power of our method in the real world.

\noindent{\bf Qualitative Results.}
We also show several visualization results in Fig.~\ref{fig:vis} on both Human3.6M and MPI-INF-3DHP with two state-of-the-art methods, MHFormer and P-SMTO. It is obvious that our model achieves excellent performance with more accurate angles on these two challenging datasets.\vspace{-0.3em}

\subsection{Ablation Study}\vspace{-0.3em}
The number of input frames is an important cause influencing the accuracy. We present the results with both CPN detected and ground truth 2D poses in Table~\ref{tab:receptive}. With the increase of the frame number, the results become more accurate.

To further explore the effectiveness of components in our model, we conduct an additional ablation study on Human3.6M under MPJPE. Considering the time efficiency, we choose the number of 27 frames. As the results shown in Table~\ref{tab:ablation}, the structural features derived from the kinematic cues in SPR and enhanced in STE improve the accuracy by 9.4mm, while the reuse of estimated results can refine the performance to 55.5mm without the help of human priors. When the structural features get enhanced and work with RR to interact with estimated results iteratively, the performance is improved by a large margin of 11.5mm to 45.8mm. These indicate that the structural features effectively introduce human priors and can be better utilized by the recursive pipeline in RR to further improve the performance.

\section{Conclusion}
\label{sec:conclusion}
In this paper, we propose EvoPose, a novel recursive transformer for 3D HPE with kinematic structure priors. EvoPose first extracts structural features from the kinematic tree to introduce human priors and builds interactions between these features and 2D pose sequences to estimate 3D poses. Then EvoPose reuses previous estimations combined with the structural features to further utilize human priors for refinement. Extensive experiments show that the proposed EvoPose has a large performance advantage in real-world scenarios.

\bibliographystyle{IEEEbib}
\bibliography{strings,refs}

\end{document}